\documentclass{article}
\usepackage{arxiv}                  

\usepackage[utf8]{inputenc}
\usepackage[T1]{fontenc}
\usepackage{microtype}
\usepackage{amsmath,amssymb}
\usepackage{graphicx}
\usepackage[table]{xcolor}          
\usepackage{booktabs}
\usepackage{tabularx}
\usepackage{threeparttable}
\usepackage[font=small,labelfont=bf,labelsep=period]{caption}
\usepackage{authblk}
\usepackage{etoolbox}
\usepackage[numbers,sort&compress]{natbib}
\usepackage{hyperref}
\usepackage{xurl}

\definecolor{natDeep}{HTML}{1F4E79}
\definecolor{natTint}{HTML}{EEF3F8}
\definecolor{natRule}{HTML}{1F4E79}
\arrayrulecolor{natRule}
\newcommand{\hd}[1]{\textcolor{white}{\textbf{#1}}}                       
\newcommand{\grp}[1]{\multicolumn{5}{l}{\rule{0pt}{2.6ex}\textbf{\textit{#1}}}\\} 
\newcommand{\Tfull}{T_{\mathrm{full}}}
\newcommand{\Tred}{T_{\mathrm{red}}}

\hypersetup{
  colorlinks=true, linkcolor=natDeep, citecolor=natDeep, urlcolor=natDeep,
  pdftitle={What Language Adds, and at What Cost: Four Controlled Probes of Pathology Vision-Language Models},
  pdfauthor={Wenhao Zhang et al.},
  pdfsubject={cs.CV; eess.IV},
  pdfkeywords={computational pathology, vision-language models, evaluation, visual grounding, robustness, survival analysis}
}

\renewcommand{\shorttitle}{What Language Adds, and at What Cost}
\renewcommand{\headeright}{A Preprint}
\renewcommand{\undertitle}{}                                
\makeatletter
\patchcmd{\@maketitle}{\textsc{\undertitle}\\}{}{}{}        
\makeatother

\title{What Language Adds, and at What Cost:\\
Four Controlled Probes of Pathology Vision--Language Models}

\author[1]{Wenhao Zhang}
\author[2]{Zhongliang Zhou}
\author[1]{Yi Ping}
\author[1]{Zirui Wang}
\author[1]{Sheng~Li\thanks{Corresponding author: \href{mailto:vga8uf@virginia.edu}{vga8uf@virginia.edu}}}
\affil[1]{School of Data Science, University of Virginia, Charlottesville, VA 22903, USA}
\affil[2]{Merck \& Co., Inc., Rahway, NJ 07065, USA}
\date{}

\begin{document}
\maketitle

\begin{abstract}
Vision--language models (VLMs) promise to make computational pathology queryable and able to draft reports, but how much language adds beyond the vision backbone is unclear, because almost every system is evaluated by its developers on test sets of their own choosing. We grade published evidence by how it was produced and test four claims with controlled probes that reuse released checkpoints and public data, varying one input at a time. Report text adds prognostic signal: adding a frozen embedding of the stage-redacted report to a slide embedding raises the pooled C-index by 0.019 for progression-free interval and by 0.016 for overall survival across 1{,}633 TCGA patients, and restoring the stage statement adds nothing. Prompts add no measurable accuracy at inference: with features and pooling fixed, zero-shot classification is statistically indistinguishable from a supervised linear probe for lung and renal cancer subtyping. Slide-level question answering is barely grounded: SlideChat answers 74.82\% of SlideBench-VQA(TCGA) items correctly with the slide and 74.41\% without it, whereas its language backbone alone reaches 43.52\%. Paraphrasing leaves accuracy unchanged but flips 8--23\% of answers. Self-reported gains from agentic navigation and few-shot language priors cannot yet be attributed to language. Language thus adds real but narrow value, and we propose minimum controls that would expose these failure modes in future evaluations.
\end{abstract}


\section{Introduction}\label{sec:intro}

Vision--language models (VLMs) couple a histology encoder to a text encoder or a large language model, so that one system can classify tissue from a prompt, retrieve similar cases, answer free-text questions and draft reports \citep{huang2023plip,lu2024conch,lu2024pathchat,ding2025titan,xiang2025musk}. Pure-vision models already reach clinical-grade accuracy on narrow tasks \citep{campanella2019clinical}, so the question for adoption is whether language makes pathology AI better or merely more persuasive: free text can assert findings that are not in the tissue, a fluent rationale can pass for evidence, and an answer that changes with the wording of a question is incompatible with diagnosis.

Published results cannot settle this. Almost every system is evaluated by the group that built it, on test data of its own choosing and under its own protocol; the most-used benchmarks are exposed to contamination (Section~\ref{sec:bench-landscape}); and the decisive comparisons are rarely run: whether language improves subtyping and survival, where vision is already strong, whether slide-level answers depend on the slide, and whether they survive rewording.

Here we weigh these claims directly. We grade evidence by how it was produced and run four controlled probes under one leakage-controlled harness, reusing released checkpoints and public data and varying a single input at a time. The probes isolate the language increment on survival prediction (Probe~1) and subtyping (Probe~2), test whether slide-level visual question answering (VQA) is grounded in the slide (Probe~3), and measure robustness to semantically equivalent rewording (Probe~4). We then grade the self-reported evidence for agentic navigation and few-shot language priors and distil the results into minimum controls. This article expands the evidence-weighing analysis of our companion review of 51 pathology VLMs \citep{zhang2026review}, from which literature-derived statements are drawn; following that review, we distinguish contrastive encoders, instruction-tuned assistants, reasoning-tuned models, agents and VLM-augmented multiple-instance learning (MIL).

\section{Evidence standards and benchmarks}\label{sec:bench-landscape}

We weigh each published result by how it was produced, in three tiers: an independent re-evaluation of several systems under one protocol (tier~1), a third-party reproduction of a single system (tier~2), and a number a system reports about itself (tier~3). Our probes are tier-1 evidence, as are PathVLM-Eval \citep{gilal2025pathvlm} and the feature-extractor benchmark of \citet{neidlinger2025benchmarking}.

Table~\ref{tab:benchmarks} lists the public benchmarks in current use. Most comparisons rest on three patch-level sets, PathMMU, PathVQA and Quilt-VQA, and all three are exposed to contamination: their images come from the literature, textbooks, educational videos and social media, the kinds of source that also populate pathology pretraining corpora \citep{huang2023plip,ikezogwo2023quilt1m}, and their question--answer pairs were generated semi-automatically or with language-model assistance. Slide-level benchmarks built from TCGA typically share their source cohort with the instruction-tuning data of the assistants evaluated on them \citep{chen2025slidechat,liang2024wsillava}. We therefore anchor the analysis on leakage-controlled evaluations.

\begin{table}[tbp]
\centering\footnotesize
\setlength{\tabcolsep}{4pt}
\renewcommand{\arraystretch}{1.18}
\begin{threeparttable}
\caption{Public evaluation benchmarks used by current pathology VLMs, with counts as reported by their sources. Items count images, whole-slide images (WSIs), patients or pairs; Questions count question--answer pairs. Gran., granularity (G1, patch; G2, region of interest; G3, whole slide); MCQ, multiple-choice question; VQA, visual question answering; ST, spatial transcriptomics; NR, not reported; ---, not applicable. WSI-Bench counts include its training split.}
\label{tab:benchmarks}
\begin{tabularx}{\linewidth}{>{\raggedright\arraybackslash}p{4.4cm}
                             >{\raggedright\arraybackslash}p{2.3cm}
                             >{\raggedright\arraybackslash}p{1.8cm}
                             c
                             >{\raggedright\arraybackslash}X}
\specialrule{1.1pt}{0pt}{0pt}
\rowcolor{natDeep}
\hd{Benchmark} & \hd{Items} & \hd{Questions} & \hd{Gran.} & \hd{Task type}\\
\specialrule{0.7pt}{0pt}{0pt}
\grp{Patch-level reasoning and VQA}
PathMMU \citep{sun2024pathmmu}                          & 24{,}067         & 33{,}428       & G1    & MCQ diagnostic reasoning\\
\rowcolor{natTint}
PathVQA \citep{he2020pathvqa}                           & 4{,}998          & 32{,}799       & G1    & Open- and closed-ended VQA\\
Quilt-VQA \citep{seyfioglu2024quiltllava}               & 985              & 1{,}283        & G1    & Open- and closed-ended VQA\\
\rowcolor{natTint}
PathQABench-Public \citep{lu2024pathchat}               & TCGA subset      & Expert-curated & G1/G2 & Diagnostic MCQ\\
\grp{ROI- and WSI-level VQA}
PathMMU-HR\textsuperscript{2} \citep{sun2025cpathagent} & NR               & 1{,}668        & G2    & VQA on $16{,}000^{2}$\,px regions\\
\rowcolor{natTint}
WSI-VQA \citep{chen2024wsivqa}                          & 977 WSIs         & 8{,}672        & G3    & Slide-level VQA\\
WSI-Bench \citep{liang2024wsillava}                     & 9{,}850 WSIs     & 179{,}569      & G3    & VQA ($3\times11$ task grid)\\
\rowcolor{natTint}
SlideBench-VQA (TCGA) \citep{chen2025slidechat}         & 732 patients     & 7{,}827        & G3    & VQA (13 tasks)\\
SlideBench-VQA (BCNB) \citep{chen2025slidechat}         & 1{,}058 patients & 7{,}274        & G3    & VQA (zero-shot)\\
\grp{Report generation and specialised tasks}
PathText \citep{chen2024wsicaption}                     & 9{,}009 pairs    & ---            & G3    & WSI-to-text report generation\\
\rowcolor{natTint}
REG2025 \citep{reg2025}                                 & 10{,}494 pairs   & ---            & G3    & Report generation, 7 organs\\
HEST-Bench \citep{jaume2024hest}                        & ST-paired WSIs   & ---            & G2/G3 & Gene-expression prediction\\
\rowcolor{natTint}
PathGLS \citep{chen2026pathgls}                         & NR               & ---            & G1/G3 & Reference-free factuality\\
\specialrule{1.1pt}{0pt}{0pt}
\end{tabularx}
\end{threeparttable}
\end{table}

\section{Controlled probes}\label{sec:methods}

All probes freeze the representation, hold the predictor and the evaluation split fixed, and vary a single input. Models are public releases used without modification, and assistants run under their authors' inference code. The encoder probes use patient- and tissue-source-site-disjoint five-fold splits, which keep site signatures from crossing folds \citep{howard2021impact}; the VQA probes are pure inference over released items. We report 95\% confidence intervals (CIs) from 1{,}000 bootstrap resamples at a fixed seed, resampling cases in Probes~1--2 and questions in Probes~3--4, and call an effect significant when its CI excludes zero.

\paragraph{Probe 1: survival.} Slides from TCGA-BRCA, -LUAD and -COAD \citep{weinstein2013tcga} are paired with progression-free interval (PFI) and overall survival (OS) from the TCGA Clinical Data Resource \citep{liu2018tcgacdr} (pooled $n=1{,}633$; 394 PFI and 345 OS events). Report text is the report-derived text of the public SlideInstruct/SlideBench-Caption release \citep{chen2025slidechat}, filtered to remove follow-up and outcome sentences. A frozen PubMedBERT report embedding \citep{gu2021pubmedbert} is concatenated with the frozen TITAN slide embedding \citep{ding2025titan}, and an L2-penalised linear Cox model \citep{cox1972regression} is fitted. Six arms differ only in their input: A, vision; B, vision with the full report; $B'$, vision with a stage-redacted report; C, structured stage and grade; $\Tfull$ and $\Tred$, report text alone. Pooled out-of-fold Harrell's C-index \citep{harrell1982evaluating} is compared through four contrasts: clean language increment ($B'-A$), staging leakage ($B-B'$), text-channel leakage ($\Tfull-\Tred$) and text versus vision ($\Tred-A$).

\paragraph{Probe 2: subtyping.} TCGA NSCLC (LUAD versus LUSC; $N=930$) and RCC (KIRC, KIRP and KICH; $N=718$) slides are embedded by three language-aligned encoders, CONCH v1 \citep{lu2024conch}, CONCH v1.5 and the TITAN slide encoder built on it \citep{ding2025titan}, and by the pure-vision Prov-GigaPath \citep{xu2024gigapath}. Each yields one vector per slide, 
on which an L2-penalised multinomial logistic head is fitted and scored by balanced accuracy. Within CONCH v1, zero-shot classification with top-$k$ or mean pooling over tiles is compared with supervised linear probes on the same features.

\paragraph{Probe 3: grounding.} SlideChat \citep{chen2025slidechat} and WSI-LLaVA \citep{liang2024wsillava}, with two text-only models, Qwen2.5-7B-Instruct \citep{qwen2024qwen25} (the language backbone of SlideChat) and Qwen3-8B \citep{qwen2025qwen3}, answer the 1{,}227 SlideBench-VQA(TCGA) multiple-choice items whose slides could be re-encoded locally (CLAM tiling \citep{lu2021clam}, CONCH features).
Relative to the unperturbed condition (P0), we remove the slide, zero its features, substitute a mismatched slide, cyclically shift the options, or replace the correct option with ``none of the above'' (NOTA). A single deterministic parser reads the answers; the most-frequent-letter baseline is 40.7\%.

\paragraph{Probe 4: wording.} SlideChat, Qwen3-VL \citep{qwen2025qwen3vl}, InternVL3 \citep{zhu2025internvl3} and a text-only Qwen3 answer a matched multiple-choice subset of SlideBench-VQA (TCGA and BCNB). With slide, question content and correct answer fixed, each item is posed canonically and in banks of semantically equivalent paraphrases, of prompt reframings (a different framing sentence or an added persona line), and of both. We report the accuracy change (canonical minus perturbed), the fraction of items whose answer flips, and answer consistency across each bank.

Linear heads and bootstrap run on CPU; zero-shot and multimodal inference run on up to two NVIDIA H200 NVL GPUs in bf16, and no model is trained or fine-tuned. Claims that the probes do not address are graded from the 51 systems of the companion review \citep{zhang2026review}.

\section{Results}\label{sec:findings}

\begin{figure}[tbp]
\centering
\includegraphics[width=0.82\linewidth]{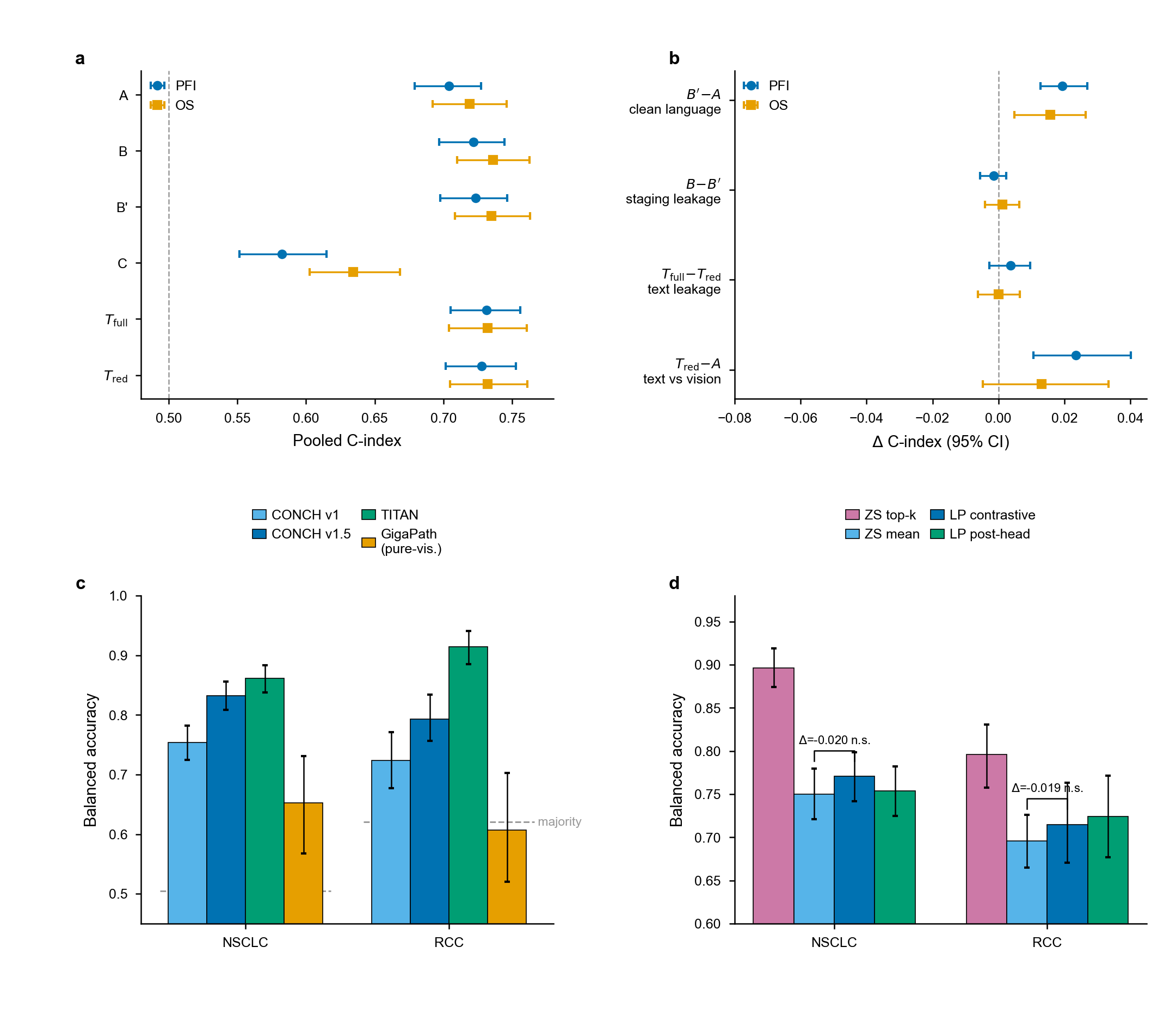}
\caption{\textbf{Probes 1 and 2: language adds prognostic signal but no discriminative power at inference.}
\textbf{(a)}~Pooled out-of-fold Harrell's C-index for the six survival arms (A, vision; B, vision${+}$full report; $B'$, vision${+}$stage-redacted report; C, structured stage and grade; $\Tfull$, $\Tred$, report text alone) for PFI and OS ($n=1{,}633$; 394 PFI and 345 OS events).
\textbf{(b)}~Survival contrasts: clean language increment ($B'-A$), staging leakage ($B-B'$), text-channel leakage ($\Tfull-\Tred$) and text versus vision ($\Tred-A$).
\textbf{(c)}~Balanced accuracy of supervised probes for NSCLC ($N=930$) and RCC ($N=718$) subtyping; dashed lines, majority baselines.
\textbf{(d)}~Within CONCH v1: zero-shot classification with top-$k$ or mean pooling, and linear probes; brackets, mean-pooled zero-shot minus contrastive-space linear probe.
Error bars, 95\% bootstrap CIs.}
\label{fig:p1-probes}
\end{figure}

\subsection{Report text adds prognostic signal that survives stage redaction}\label{sec:attribution}

Vision--language foundation models report better survival prediction than pure-vision models \citep{ding2025titan,xiang2025musk}, but reports state tumour stage, so the gain could merely restate it. The clean increment $B'-A$ was positive and significant in all six cohort--endpoint settings 
and pooled to $+0.019$ C-index for PFI and $+0.016$ for OS (Fig.~\ref{fig:p1-probes}b). Stage does not explain it: both leakage contrasts were null, the redacted report alone was at least as prognostic as the slide embedding, and structured stage and grade (arm~C) formed the weakest arm (Fig.~\ref{fig:p1-probes}a). The increment is therefore not a restatement of the stage statement. It is information in the report that complements the slide, captured by a frozen text encoder rather than by a VLM, and redaction leaves stage-correlated findings such as tumour size in place.

\subsection{Language pays through pretrained features, not prompts}\label{sec:supervised-probe}

Language enters a contrastive VLM twice, as pretraining supervision and as zero-shot prompts; if predictive power lives in the features, a laboratory can replace prompts with a supervised classifier and avoid prompt sensitivity. Under identical supervised probes, all three language-aligned encoders outperformed Prov-GigaPath on both tasks (Fig.~\ref{fig:p1-probes}c). A probe uses no prompt, so the advantage resides in the features; but the encoders also differ in backbone, data, dimension and aggregation, so it cannot be attributed to language. The within-encoder arm isolates the prompt: with CONCH v1 features and pooling fixed, zero-shot classification was statistically indistinguishable from the linear probe (balanced-accuracy difference $-0.020$ for NSCLC and $-0.019$ for RCC; Fig.~\ref{fig:p1-probes}d). Top-$k$ zero-shot classification scored higher, but its margin vanished under shared pooling, an instance-selection rather than a prompt effect. With CONCH ranked first of 19 foundation models as a weakly supervised feature extractor in an independent benchmark \citep{neidlinger2025benchmarking}, the evidence indicates that, for in-distribution subtyping, vision--language pretraining pays, if at all, through the features it shapes rather than through prompts.

\begin{figure}[tbp]
\centering
\includegraphics[width=0.95\linewidth]{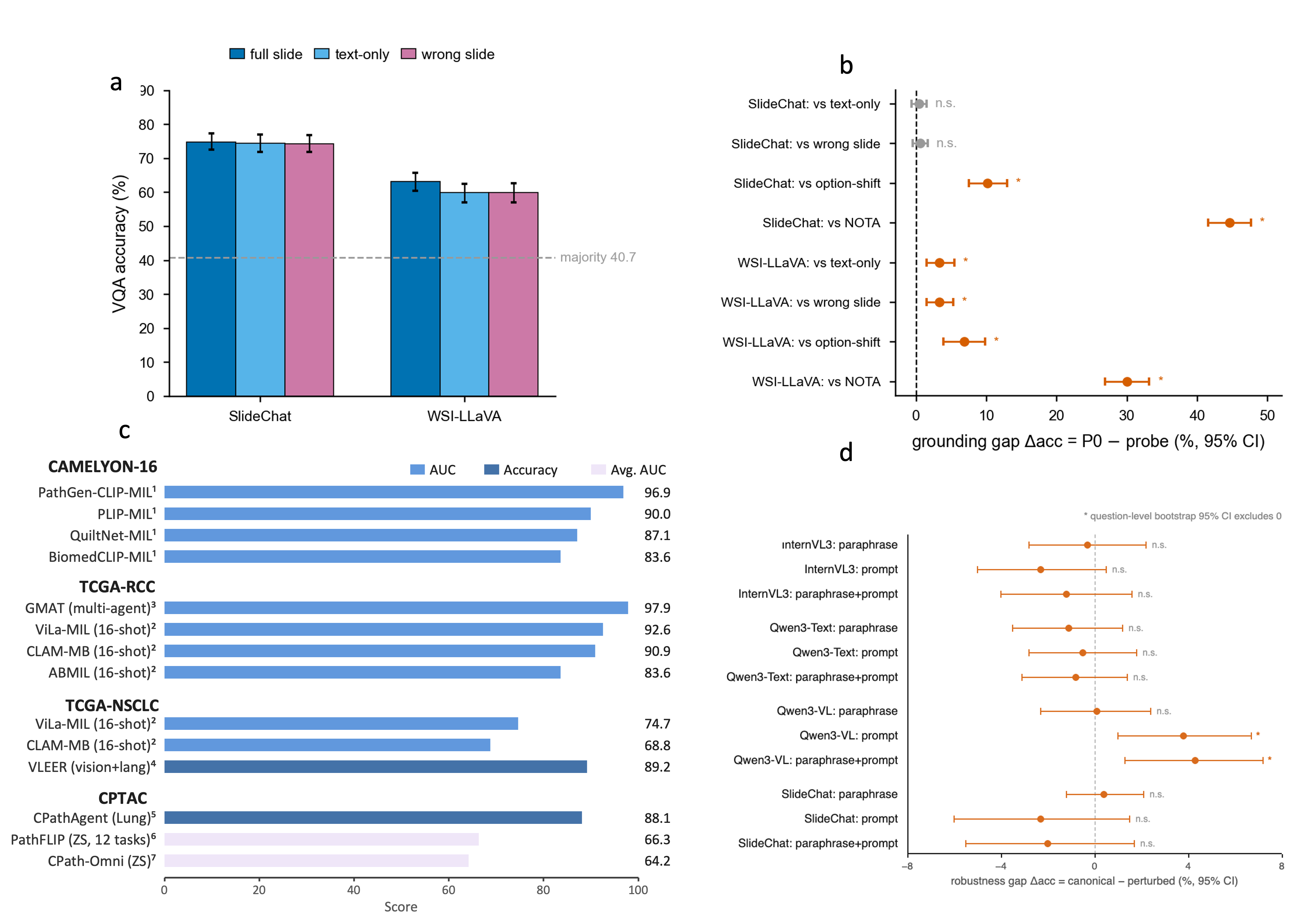}
\caption{\textbf{Probes 3 and 4, and self-reported results.}
\textbf{(a)}~SlideBench-VQA(TCGA) accuracy with the full, removed (text-only) or mismatched slide; dashed line, most-frequent-letter baseline (40.7\%).
\textbf{(b)}~Grounding gaps ($\Delta\mathrm{acc}=\mathrm{P0}-\mathrm{probe}$; P0, unperturbed condition): slide-removal and wrong-slide gaps are null (SlideChat) or small (WSI-LLaVA), whereas option shift lowers and none-of-the-above (NOTA) substitution collapses accuracy.
\textbf{(c)}~Self-reported downstream results by benchmark; shading encodes the metric, so bars compare only within a benchmark and metric. Superscripts index the reporting publication; no value was re-evaluated here.
\textbf{(d)}~Robustness gaps ($\Delta\mathrm{acc}=\text{canonical}-\text{perturbed}$) under paraphrase, prompt reframing and both.
Error bars, question-level bootstrap 95\% CIs; asterisks, CI excludes zero; n.s., not significant.}
\label{fig:p2-grounding}
\end{figure}

\subsection{Slide-level VQA accuracy does not depend on the slide}\label{sec:grounding}

A multiple-choice benchmark can be passed without looking, by exploiting answer distributions and option positions. SlideChat answered 74.82\% of items correctly with the slide, and its text-only and mismatched-slide gaps were indistinguishable from zero, including on grading and staging questions; WSI-LLaVA lost a few points, significantly (Fig.~\ref{fig:p2-grounding}a,b). The prior is learned in training: Qwen2.5-7B, SlideChat's own backbone, answered 43.52\% correctly without the slide, close to the 40.7\% letter baseline, whereas SlideChat, equally blind, answered 74.41\%, so about 31 points come from regularities of its instruction-tuning data rather than from the slide. The assistants were instead sensitive to answer format: shifting option positions lowered accuracy significantly, and substituting NOTA for the correct option pushed SlideChat below the letter baseline, so neither model rejects the remaining options once the correct content is removed.

\subsection{Rewording leaves accuracy intact but changes the answers}\label{sec:robustness}

Under paraphrase alone, the accuracy change was indistinguishable from zero for all four systems (Fig.~\ref{fig:p2-grounding}d), so we detect no accuracy cost. The answers, however, were unstable: 8--23\% of items changed answer under paraphrase and 25\% under paraphrase plus reframing, where consistency fell to 0.57. Even SlideChat, the most stable system, changed a quarter of its answers when only the framing sentence varied, and a single persona line shifted Qwen3-VL's accuracy significantly, the only accuracy changes whose intervals excluded zero. Leaderboards cannot see this instability; in the clinic, where the same case must yield the same diagnosis, it is disqualifying.

\subsection{Self-reported gains cannot yet be attributed}\label{sec:val-p4}

Two further claims rest on self-reported evidence (tier~3; Fig.~\ref{fig:p2-grounding}c). Agentic navigation produces its largest margins where they are least separable. CPathAgent reaches 88.6\% on PathMMU-HR\textsuperscript{2}, 15.4 points above Gemini-2.5-Pro answering in a single pass, yet at slide level it leads ABMIL by 2.9 points on average and trails TITAN on CPTAC-Lung (88.1\% versus 91.9\%) \citep{sun2025cpathagent}. The margin thus depends on scale, and only part of it is navigation: prompted to navigate in the same way, Gemini-2.5-Pro rises from 73.2\% to 76.4\%. With no intervals or seeds reported, neither component can be tested. Language priors in MIL earn smaller and more consistent margins: ViLa-MIL, VLEER and GMAT report few-shot gains over language-free aggregators while training fewer than $10^{7}$ parameters \citep{shi2024vilamil,nguyen2025vleer,quang2025gmat}, for example 1.7 and 5.9 AUC points for ViLa-MIL over CLAM-MB with 16 shots (Fig.~\ref{fig:p2-grounding}c). Every such comparison comes from the proposing paper without independent or leakage-controlled replication, so we grade this evidence as weakly supported.

\section{Discussion}\label{sec:discussion}

\paragraph{What accuracy does not capture.} Visibility reorders clinical risk: the hallucinations of generative systems surface as text that a reader can check, whereas score-only systems (contrastive encoders and VLM-augmented MIL) fail through correlational shortcuts, such as a frozen encoder passing a tissue-source-site signature onward \citep{howard2021impact}, and leave a confident, wrong prediction with no trace. Long-tail behaviour is rarely reported: among the catalogued systems, HiPath falls from 70.2\% on head terms to 2.6\% on tail terms \citep{yuan2026hipath}, MR-PLIP reaches 0.745 balanced accuracy on 30 EBRAINS brain-tumour subtypes \citep{albastaki2025mrplip}, CPathAgent scores 64.3\% on three-class BRACS regions \citep{sun2025cpathagent}, and paediatric tumours are assessed only by KEEP, on 816 WSIs from one hospital \citep{zhou2026keep}. Calibration and abstention are absent: networks are poorly calibrated by default \citep{guo2017calibration}, free text carries no calibrated confidence, selective prediction \citep{geifman2017selective} goes unused, and none of the 51 systems reports calibrated confidence. Compute is reported sporadically and in incompatible units: a few papers give training budgets \citep{li2023llavamed,yuan2026hipath}, only LAMMI-Pathology reports a memory footprint \citep{su2026lammi}, and none compares runtimes across paradigms or estimates carbon cost \citep{lacoste2019cocodecarb}, although instruction-tuned assistants (7B--15B parameters) need roughly $10^{4}$ visual tokens and 40--80\,GB of GPU memory per WSI at inference.

\paragraph{Minimum controls.} Each failure reported here was exposed by a cheap control. We propose that evaluations of pathology VLMs report, at a minimum: (i)~for slide-conditioned VQA, text-only, mismatched-slide, option-shift and NOTA arms alongside the majority-answer baseline; (ii)~answer-flip rates and consistency under paraphrase and reframing; (iii)~stage-redacted and text-only arms for claims that text improves prognosis; (iv)~within-encoder comparisons for claims about prompts; (v)~CIs and seeds for every headline comparison, preferably with independent replication; and (vi)~compute in six fields: GPU type and count, GPU-hours, peak training memory, peak inference memory, visual tokens per WSI, and end-to-end latency on a reference slide. For design, language at inference is best spent where it is irreplaceable (prognosis, interaction and reporting), and grounding should become a training objective: answers should cite tissue evidence, change when it changes, abstain when it is absent and stay invariant to paraphrase. None of this replaces clinical validation through multisite testing, subgroup reporting and prospective reader studies.

\paragraph{Limitations.} The probes cover released checkpoints on TCGA-derived data, two slide-level assistants, three language-aligned encoders and one pure-vision encoder; reasoning-tuned, agentic and MIL systems are not probed directly. In Probe~1, a null leakage contrast would also arise if the frozen text encoder were insensitive to stage statements, no arm combines vision with structured stage and grade, 
and the report text is a processed release. In Probe~2, the cross-encoder comparison is confounded, and zero-shot and supervised heads are compared under mean pooling only, without learned instance selection such as attention-based MIL. Probes~3 and~4 use multiple-choice items, so free-text generation may behave differently, and the navigation and few-shot grades rest on self-reported numbers.

\section{Conclusion}\label{sec:conclusion}

Weighed by controlled probes rather than headline numbers, language adds real but narrow value to pathology VLMs. Report text carries prognostic signal that survives stage redaction, but prompts add no discriminative accuracy at inference, and slide-level answers are neither grounded in the slide nor stable under rewording. Until grounding, paraphrase consistency and calibration are reported as routinely as accuracy, benchmark scores should be read as upper bounds on what a model reads from the slide, not as evidence that it does.

\section*{Data and code availability}
All data are public and de-identified, and no new human-subjects data were collected. Whole-slide images and clinical endpoints are from TCGA and the TCGA Clinical Data Resource via the NIH Genomic Data Commons (\url{https://portal.gdc.cancer.gov}); report-derived text and VQA items are from the public SlideChat release (\url{https://huggingface.co/datasets/General-Medical-AI/SlideChat}). All checkpoints (CONCH v1 and v1.5, TITAN, Prov-GigaPath, PubMedBERT, SlideChat, WSI-LLaVA, Qwen2.5-7B-Instruct, Qwen3-8B, Qwen3-VL and InternVL3) are public releases. A maintained list of pathology VLMs is available at \url{https://github.com/wenhaozhang0066/Awesome-Pathology-VLMs}.

\section*{Author contributions}
W.Z., Z.Z. and S.L. conceived the project. W.Z., Y.P. and Z.W. performed the experiments. W.Z., Y.P., Z.W., P.W., H.Y. and T.W. wrote the manuscript. W.Z., Z.Z., Q.Z., Y.Y., J.K., E.T.R., J.B.W., C.E.P. and S.L. revised the manuscript. Z.Z., J.K., E.T.R., J.B.W., C.E.P. and S.L. supervised the project. All authors read and approved the final manuscript.

\section*{Competing interests}
Z.Z., Q.Z., J.K., E.T.R., J.B.W. and C.E.P. are employees of Merck Sharp \& Dohme LLC, a subsidiary of Merck \& Co., Inc., Rahway, NJ, USA, and may hold stock in Merck \& Co., Inc. The remaining authors declare no competing interests.


\bibliographystyle{unsrtnat}
\bibliography{refs}

\end{document}